\documentclass[letterpaper, 10 pt, conference]{ieeeconf}
\IEEEoverridecommandlockouts                              % This command is only needed if
\usepackage{url}
\usepackage{latexsym}
\usepackage{subfigure}
\usepackage{float}
\usepackage{graphicx}
\usepackage{color}
\usepackage{cite}
\usepackage{listings}
\usepackage{mathrsfs}
\usepackage{amssymb}
\usepackage{pstool}
\usepackage[ruled]{algorithm2e}
\usepackage{multirow}
\usepackage{amsmath}
{

}

\makeatletter
\renewcommand{\@IEEEsectpunct}{}
\makeatother

\usepackage{latexsym}
\usepackage{subfigure}
\usepackage{graphicx}
\usepackage{color}
\usepackage{amsmath}
\usepackage{mathtools}
\usepackage{booktabs}

\newcommand{\figref}[1]{Fig.~\ref{fig:#1}}

\newcommand{\tabref}[1]{Table~\ref{tab:#1}}

\renewcommand{\eqref}[1]{(\ref{eq:#1})}

\newcommand{\mtrx}[1]{\mathrm{\mathbf{#1}}}

\def\sign{\mathop{\rm sign}\nolimits}

\def\atan2{\mathop{\rm atan2}\nolimits}
\newcommand{\bs}[1]{\ensuremath{{\boldsymbol{#1}}}}
\def\atan2{\mathrm{atan2}}

\title{\LARGE \bf Motion Planning and Control of an Overactuated 4-Wheel Drive with Constrained Independent Steering}%PN

\author{
  Shiyu Liu, Ilija Had\v{z}i\'{c}, Akshay Gupta, and Aliasghar Arab%
  \thanks{\scriptsize The work reported in the paper was performed in its entirety while Liu was affiliated with Nokia Bell Labs, France and Arab was affiliated with Nokia Bell Labs, USA. Had\v{z}i\'{c} and Gupta are with Nokia Bell Labs, USA and Arab is now with New York University, Tandon School of Engineering, Mechanical and Aerospace Engineering Department.
      }
}

\begin{document}
\maketitle

\begin{abstract}
  \label{abstract}
  This paper addresses motion planning and control of an overactuated
  4-wheel drive train with independent steering (4WIS) where mechanical
  constraints prevent the wheels from executing full 360-degree rotations
  (swerve).
  The configuration space of such a robot is constrained and contains
  discontinuities that affect the smoothness of the robot motion.
  We introduce a mathematical formulation of the steering constraints
  and derive discontinuity planes that partition the velocity space into
  regions of smooth and efficient motion.
  We further design the motion planner for path tracking and obstacle
  avoidance that explicitly accounts for swerve constraints and the
  velocity transition smoothness.
  The motion controller uses local feedback to generate actuation
  from the desired
  velocity, while properly handling the discontinuity crossing by temporarily
  stopping the motion and repositioning the wheels.
  We implement the proposed motion planner as an extension to ROS
  Navigation package and evaluate the system in simulation and on a physical
  robot.
\end{abstract}

%\begin{IEEEkeywords}
%  TBD
%\end{IEEEkeywords}

\section{Introduction}
\label{Intro}

Four-wheel drive with independent steering (4WIS) is a mobility mechanism
in which each wheel is actuated by two dedicated motors --- one for steering
and one for driving --- enabling independent steering and drive.
This drivetrain can theoretically achieve three degrees of freedom (3-DOF)
with zero side-slip forces~\cite{selekwa2011path}, which enables
low tire wear, energy efficiency, and smooth, versatile maneuvers.
Kinematic and dynamic control methods for this drivetrain
have been extensively studied~\cite{hang2017robust, hang2018design, yin2010robust, xu2019improving, lee2015kinematics}. All reported work
assumes unconstrained steering, that is, the ability for the wheel to
perform a full 360-degree swerve.

% TODO: cite swerve module
Although simple mechanical construction
of a full-swerve wheel exists\cite{Khairnar2023}, very often
chassis design or cabling constraints limit
the rotation range of the steering shaft~\cite{zoox, ottonomy, naio}.
These systems typically sidestep the issue by operating in
three discrete modes: longitudinal or lateral Ackermann steering,
along with an option for in-place rotation.

In this paper, we propose a cascade controller that unifies constrained
and unconstrained case.
Steering constraints are expressed through so-called {\em discontinuity planes},
which dissolve for unconstrained case. We integrate the controller with
the Robot Operating System (ROS) Navigation
package\cite{ros_navigation}, which is a commonly used software platform
for many research and industrial robots.
Although it is possible to formulate the steering constraints in a
dynamic model,
and design the controller based on more sophisticated algorithms such as
MPC~\cite{liu2022mpc, arab2021safe}, we opt for a cascaded architecture
to fit the architecture of ROS Navigation stack. In this architecture,
the global planner generates the trajectory as a set of waypoints, the local
planner selects the optimal velocity taking into account kinematic and
dynamic constraints, and the motion controller produces the actuation
that satisfies steering constraints.
The baseline local planner in ROS Navigation package is derived from
Dynamic Window
Approach (DWA)\cite{fox1997dynamic} planner, which we also use as the baseline
for our system.

\section{System Architecture}
\label{sec:probstat}

\begin{figure*}[th!]
        \vspace{1mm}
        \centering
        \subfigure[]{
          \centering
          \includegraphics[width=3.3in]{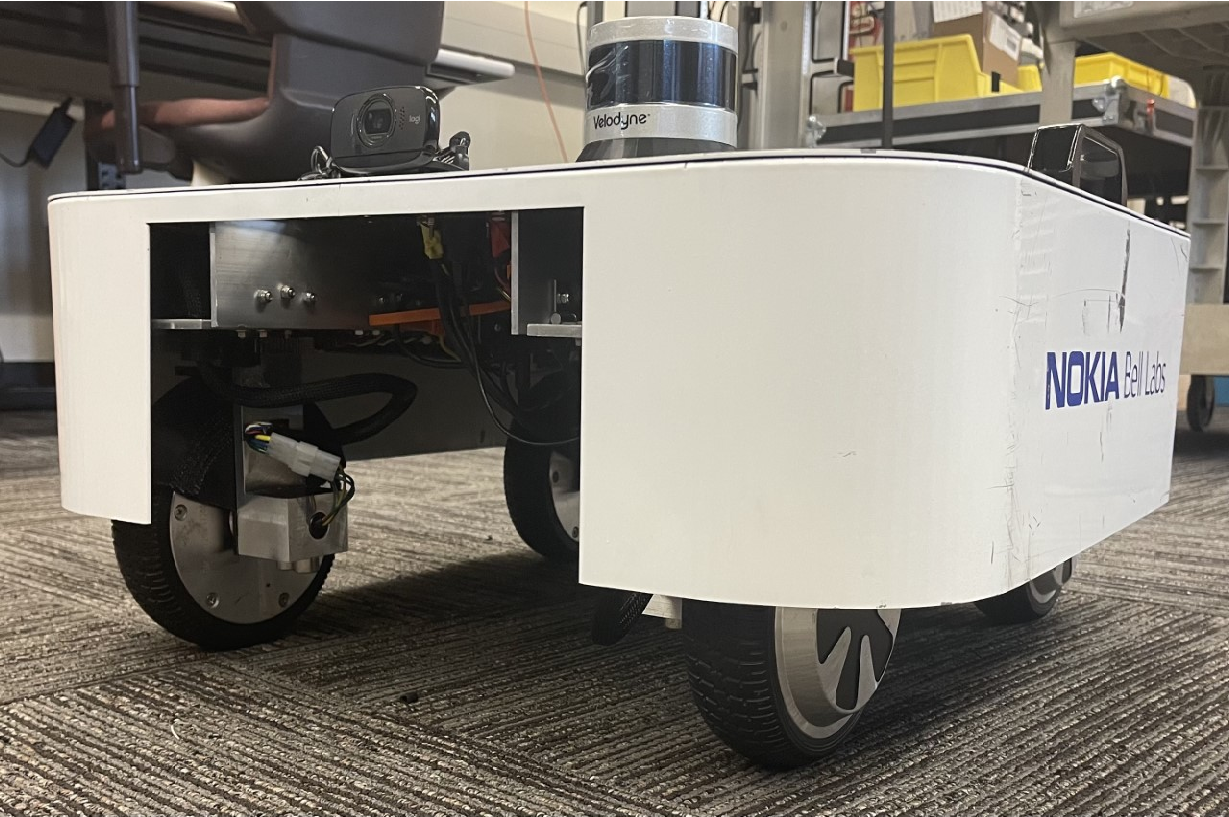}
          \label{fig:robot}
        }
        \subfigure[]{
          \vspace{-4mm}
          \centering
          \includegraphics[width=2.3in]{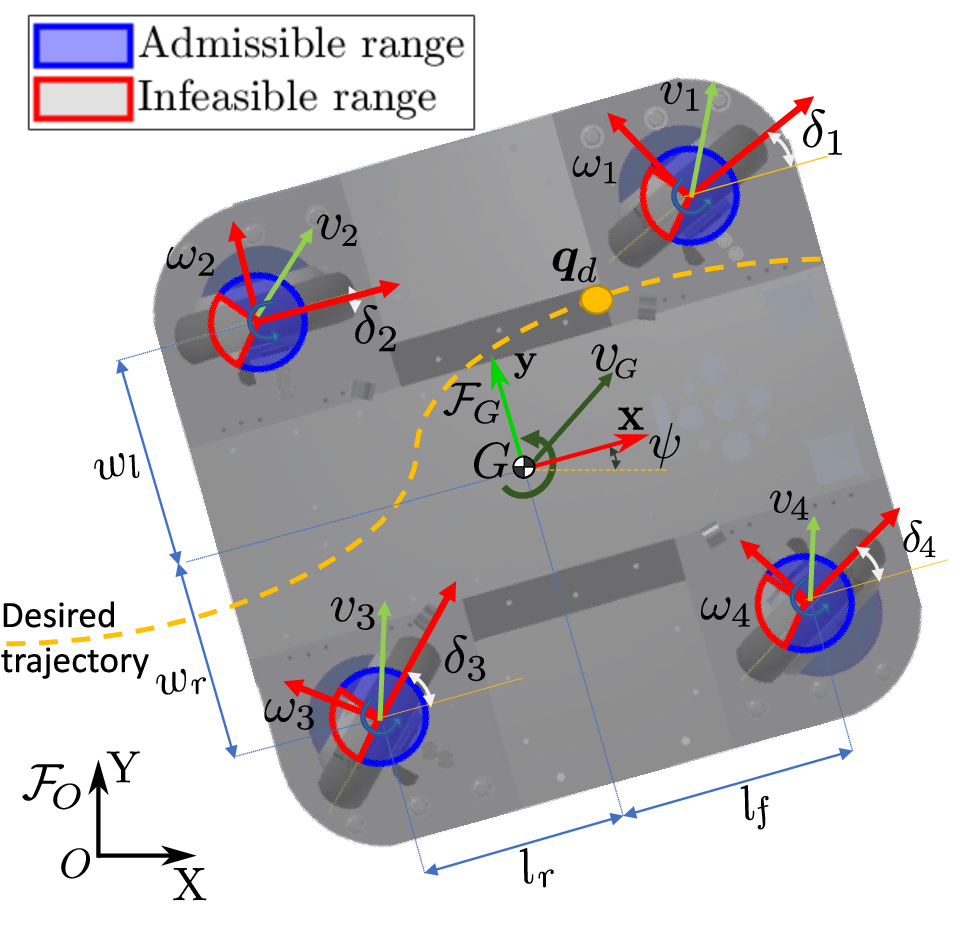}
          \label{fig:TopView}
        }
        \caption{
          A 4-wheel drive with independent steering (4WIS) robot:
          (a) front view of the experimental model and (b) top view
          diagram of the drivetrain.
        }
        \label{fig:robots}
\end{figure*}

Figure \ref{fig:robot} shows an in-house-built 4WIS drive that we use to develop the controller
and conduct the experiments. The hardware consists of
four hub-wheels with embedded drive-motors, each mounted on a rotary
mechanism driven by a dedicated steering motor.
The drive-motor power supply lines and encoder wires extend
from the motor controller, mounted on the chassis, to the
motor and they twist as the wheel steers. Conduits and fenders
prevent the wires from rubbing the spinning wheel, but the whole
cable-management mechanism also prevents the steering mechanism from performing
a full 360-degree swerve. It is this limitation that imposes kinematic
constraints that the controller must observe during the motion
and during transitions from one velocity to another.
Such constrained-swerve steering mechanism has been practically witnessed in
many industrial 4WIS platforms\cite{zoox, ottonomy, naio}.

Figure \ref{fig:TopView} shows the annotated top-view diagram used for modeling.
Geometric parameters $\mathrm{w}_l$, $\mathrm{w}_r$, $l_f$,
and $l_r$ define the position of the wheels in the local
frame of reference $\mathcal{F}_G({\bf x}, {\bf y})$.
Body velocity is denoted as $\bs{v}=[ v_{x} \; v_{y} \; \dot{\psi} ]^{T}$
where $v_{x}$, $v_{y}$ are linear velocity components in
$\mathcal{F}_G({\bf x}, {\bf y})$, and $\dot{\psi}$ is the angular
velocity (yaw rate). The wheel radius is denoted as $r_{\mathrm{w}}$.
The wheels' angular velocities around the rotating shaft are given
by $\bs{\omega}=\left[\omega_{1} \; \omega_{2} \; \omega_{3} \; \omega_{4} \right]^{T}$, and their
linear velocities are defined by
$\bs{v}_{\mathrm{w}}=\left[v_{1} \; v_{2} \; v_{3} \; v_{4} \right]^{T}$.
The steering angles of the wheels are denoted by
$\bs{\delta} = \left[\delta_{1} \; \delta_{2} \; \delta_{3} \; \delta_{4} \right]^{T}$.
The order of all vector components with index $i \in \{1, 2, 3, 4\}$
follows the convention: front-left, rear-left, rear-right and front-right.

The objective of motion control is to deliver the sequence of
velocity vectors $\bs{v}$ in $\mathcal{F}_G({\bf x}, {\bf y})$ that
advances the robot towards the goal or the next waypoint ${\bf q}_d$ in the reference
frame $\mathcal{F}_O(\text{X}, \text{Y})$. The motion is smooth if the
generated sequence successfully avoids constraints. When such avoidance
is not possible, the robot must temporarily stop the motion and
reposition the wheels before continuing. The constraints are kinematic
in nature, but they exist regardless of whether the controller is kinematic
or dynamic.

\section{Steering Wheel Constraints}
\label{sec:phys}
We first derive the steering constraints presented above.
From 4WIS inverse kinematics \cite{arab2021safe}, it is straightforward to show
that for a given velocity vector, the wheel positions are

\begin{equation}
  \bs{\delta} = \begin{bmatrix}
   & \atan2\left(v_y+l_{\mathrm{f}}\dot{\psi},\;v_x-w_{\mathrm{l}}\dot{\psi}\right) \\
   & \atan2\left(v_y-l_{\mathrm{r}}\dot{\psi},\;v_x-w_{\mathrm{l}}\dot{\psi}\right) \\
   & \atan2\left(v_y-l_{\mathrm{r}}\dot{\psi},\;v_x+w_{\mathrm{r}}\dot{\psi}\right) \\
   & \atan2\left(v_y+l_{\mathrm{f}}\dot{\psi},\;v_x+w_{\mathrm{r}}\dot{\psi}\right)
  \end{bmatrix}.
\label{eq:calcSteer2}
\end{equation}

%As the system transitions from one state to another, the steering angles
%and wheel angular velocities change continuously. Consider a wheel
%whose steering angle $\delta_i$ is steadily increasing as part of this
%transition towards new configuration. Eventually, the steering angle
%$\delta_i$ will hit the limits $\delta_{\min}$ or $\delta_{\max}$ beyond which it cannot move
%any further. If the mechanical design allows the range of steering
%angles greater than 180 degrees (i.e., $\delta_{\min}\leq-\frac{\pi}{2}$ and $\delta_{\max}\geq%\frac{\pi}{2}$),
%then $\delta_{\min} + \pi$ or $\delta_{\max} - \pi$ must be a
%valid steering angle. Transitioning to that angle and reversing the direction
%of wheel rotation will result in a valid solution, but the actuation
%cannot change abruptly and, even if it could, it would cause an
%extreme stress to the mechanism. This scenario intuitively describes where
%the discontinuity in the configuration space comes from. We next analyze
%the problem formally.

\noindent From \figref{TopView}, steering-wheel constraints are expressed as
\begin{align}
  \delta_i \in \Omega_{\mathrm{\delta}} = \left[ \; \delta_{\min}, \delta_{\max} \; \right], \: \text{ for } i\in\{1, 2, 3, 4\},
\label{eq:steerlimit}
\end{align}

\noindent resulting in eight
inequalities (two per wheel) that
determine a set of admissible velocity configurations $\mathcal{V}$ represented by
\begin{align}
\bs{v} \in \mathcal{V} = \bigcap_{i\in \{1, 2, 3, 4\}}{\mathcal{V}_i} \subseteq \Re^{3}
\label{eq:Inequality2subset}
\end{align}

\noindent where $\mathcal{V}_i$ is defined as a subset of velocity configurations
that the robot's velocity vector $\bs{v}$ can reach without violating the steering
angle limits imposed by the $i$th wheel:

\begin{equation}
\mathcal{V}_i = \left\{\bs{v} \> | \> (\bs{b}^{(1)}_i)^T \bs{v} > 0, \> (\bs{b}^{(2)}_i)^T \bs{v} > 0 \right\}.
\label{eq:v_i}
\end{equation}
\noindent $\bs{b}^{(1)}_i, \bs{b}^{(2)}_i$
are vectors factorizing the velocity vector $\bs{v}$
to formulate inequality constraints associated with
the steering angle limits
(i.e., $\delta_i < \delta_{\max}$ and $\delta_i > \delta_{\min}$).
We derive
inequality conditions starting from \eqref{Inequality2subset} and \eqref{v_i}, considering
any given velocity configuration. First, we establish certain properties
of the steering limits.

If the steering angle limits satisfy $\delta_{\mathrm{min}} \leq -\pi$
and $\delta_{\mathrm{max}} \geq \pi$ conditions, then a solution
that satisfies~\eqref{Inequality2subset} exists for
each steering wheel $\delta_i$ in~\eqref{calcSteer2} and
for any given velocity configuration $\bs{v} \in \Re^{3}$.
Recall the property of $\operatorname{atan2}$
function, whose range is $\left(-\pi, \pi\right]$. So any $\delta_i$ computed from~\eqref{calcSteer2}
is a valid solution without crossing the steering angle limits.

If
$\delta_{\mathrm{min}} \in (-\pi, -\frac{\pi}{2}]$ and
$\delta_{\mathrm{max}} \in [\frac{\pi}{2}, \pi)$, there exists {\em at least}
one solution for any given velocity configuration $\bs{v} \in \Re^{3}$,
but some solutions may violate the steering limits, in which case,
the robot can achieve the same velocity
by switching the rotation direction
of the drive wheels and flipping the steering angles $\delta_i$:
\begin{align}
\delta_i, \omega_i = \left\{\begin{array}{lr}
\delta_i, \: \omega_i \quad & \delta_i \in \Omega_{\delta} \\
\delta_i-\pi\sign(\delta_i), \: -\omega_i \quad & \delta_i \notin \Omega_{\delta}
\end{array}.\right.
\label{eq:steerLimit2}
\end{align}
\noindent The $\sign(.)$ is the sign function and $\omega_i$ is the
rotational velocity of the $i$th drive motor.

\subsection{Discontinuity Planes}

The inequality constraints defined in \eqref{v_i},
imposed by the steering angle limits, form {\em discontinuity planes}
in the robot's velocity configuration space. These discontinuity planes
are solved by imposing the equality constraints for each wheel as
$(\bs{b}^{(1)}_i)^T \bs{v} = 0 $ and $(\bs{b}^{(2)}_i)^T \bs{v} = 0$,
for $i \in \{1, 2, 3, 4\}$.
The motion remains smooth
if the trajectory through the velocity space formed by changing velocity
does not cross any discontinuity plane.
Figure \ref{fig:discontinuityPlane} shows the discontinuity
planes for conditions discussed herein.

\subsubsection{$\delta_{\mathrm{max}}=-\delta_{\mathrm{min}}=\frac{\pi}{2}$}
\begin{figure*}[htp!]
        \vspace{1mm}
	\centering
	\subfigure[]{
		\centering
		\label{fig:pi2}
		\includegraphics[width=0.3\linewidth]{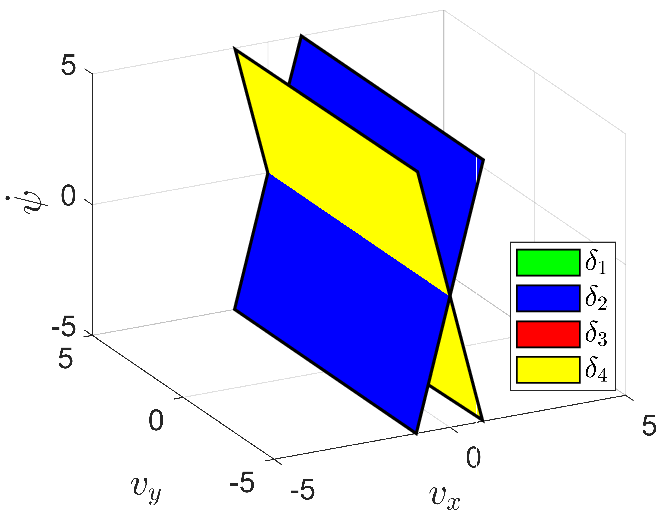}
	}
	\subfigure[]{
		\centering
		\label{fig:2pi3}
		\includegraphics[width=0.3\linewidth]{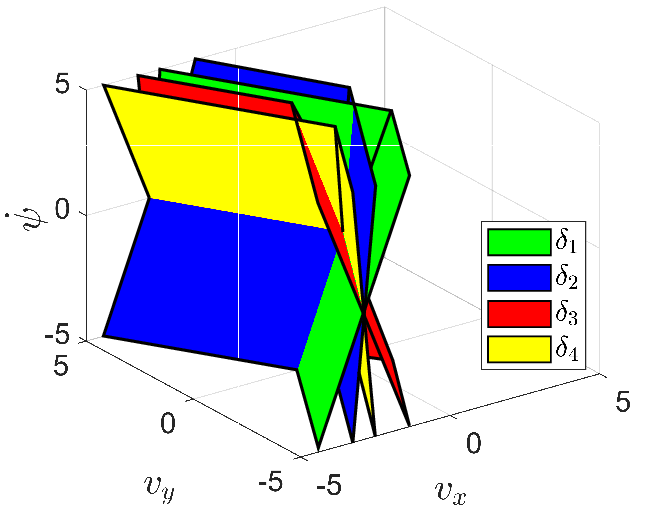}
	}
	\subfigure[]{
		\centering
		\label{fig:3pi4}
		\includegraphics[width=0.3\linewidth]{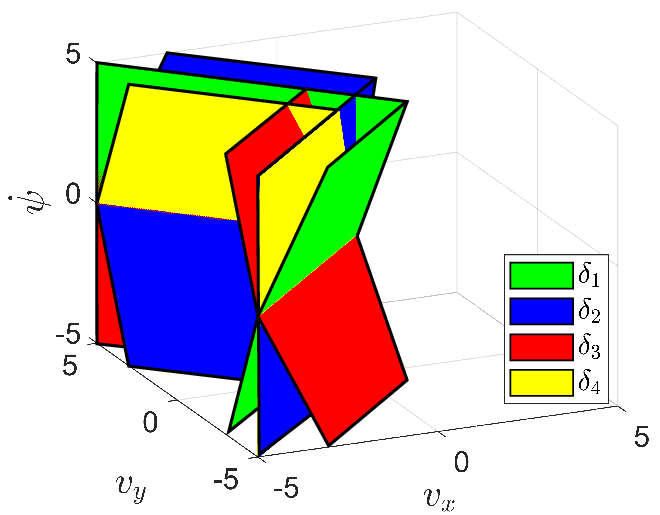}
	}
	\caption{Discontinuity planes in the $3D$ velocity configuration space.
  (a) $\Omega_{\delta}=[-\frac{\pi}{2}, \frac{\pi}{2}]$; (b) $\Omega_{\delta}=[-\frac{2\pi}{3}, \frac{2\pi}{3}]$; (c) $\Omega_{\delta}=[-\frac{3\pi}{4}, \frac{3\pi}{4}]$.}
	\label{fig:discontinuityPlane}
\end{figure*}

:
Figure \ref{fig:pi2} shows the configuration space split in four continuous regions
by two discontinuity planes. This is a special case in which eight constraints
collapse into only two planes. Specifically, when the output of
$\operatorname{atan2}$ function in \eqref{calcSteer2} is
$\pm\frac{\pi}{2}$ the discontinuity planes for four wheels reduce to
\begin{equation}
\begin{aligned}
    & {v_x-w_{\mathrm{l}}\dot{\psi}} = 0 \:\:\:\: (\text{for wheels $1$ and $2$})\\
    & {v_x+w_{\mathrm{r}}\dot{\psi}} = 0 \:\:\:\: (\text{for wheels $3$ and $4$}),
\end{aligned}
\label{eq:discontinuity_plane_pi2}
\end{equation}
that is $x=0$ for $\operatorname{atan2}(y, x) = \pm\frac{\pi}{2}$.

To avoid stopping the robot to reverse the rotational direction of the wheels, it is preferred
to keep the configuration within one of the two larger regions
of \figref{pi2}. If we further assume that forward-motion is preferred,
the constraints are
\begin{equation}
  \begin{aligned}
    & {v_x-w_{\mathrm{l}}\dot{\psi}} > 0 \\
    & {v_x+w_{\mathrm{r}}\dot{\psi}} > 0.
  \end{aligned}
  \label{eq:steer2speedpi2}
\end{equation}

\subsubsection{$\delta_{\mathrm{max}}=-\delta_{\mathrm{min}} \in (\frac{\pi}{2},\pi)$}:
Figure~\ref{fig:2pi3} and \ref{fig:3pi4} show two instances of a general case where the
configuration space is split in 12 continuous
regions by 8 discontinuity planes (i.e., two constraints per wheel). Considering
\eqref{calcSteer2} and the steering limits in \eqref{steerlimit},
the discontinuity planes corresponding to two steering constraints for the $i$-th wheel can
be given by
\begin{equation}
  \begin{aligned}
    \tan(\delta_{\mathrm{max}}) \big(v_x + w_i \dot{\psi}\big) - \big(v_y + l_i \dot{\psi}\big) & = 0 \\
    \tan(\delta_{\mathrm{min}}) \big(v_x + w_i \dot{\psi}\big) - \big(v_y + l_i \dot{\psi}\big) & = 0,
  \end{aligned}
  \label{eq:wheel_discontinuity_func}
\end{equation}
with additional constraints $v_x + w_i \dot{\psi} < 0$, where
$w_i \in \{-w_{\mathrm{l}}, -w_{\mathrm{l}}, w_{\mathrm{r}}, w_{\mathrm{r}}\}$ and
$l_i \in \{l_{\mathrm{f}}, -l_{\mathrm{r}}, -l_{\mathrm{r}}, l_{\mathrm{f}}\}$, for $i \in \{1,2,3,4\}$.
By factorizing \eqref{wheel_discontinuity_func} with respect to $\bs{v}$,
$\bs{b}^{(1)}_i$ and $\bs{b}^{(2)}_i$ can be derived as
\begin{equation}
  \begin{aligned}
    \bs{b}^{(1)}_i & = \begin{bmatrix} \tan(\delta_{\mathrm{max}}) & -1 & \big(w_i \tan(\delta_{\mathrm{max}}) - l_i\big) \end{bmatrix}\\
    \bs{b}^{(2)}_i & = \begin{bmatrix} \tan(\delta_{\mathrm{min}}) & -1 & \big(w_i \tan(\delta_{\mathrm{min}}) - l_i\big) \end{bmatrix}.
  \end{aligned}
  \label{eq:b_i}
\end{equation}
By grouping $\bs{b}^{(1)}_i$ and $\bs{b}^{(2)}_i$ for all four wheels, the 8 discontinuity planes are the solutions of $\mtrx{B}\bs{v}=\bs{0}$,
where $\mtrx{B} \in \Re^{8\times 3}$ is defined as
\begin{align}
  \mtrx{B} = \begin{bmatrix}
     -\tan(\delta_{\mathrm{max}}) &  1 &  \big(l_{\mathrm{f}} + w_{\mathrm{l}} \tan(\delta_{\mathrm{max}}) \big) \\
     -\tan(\delta_{\mathrm{max}}) &  1 & -\big(l_{\mathrm{r}} - w_{\mathrm{l}} \tan(\delta_{\mathrm{max}}) \big) \\
     -\tan(\delta_{\mathrm{max}}) &  1 & -\big(l_{\mathrm{r}} + w_{\mathrm{r}} \tan(\delta_{\mathrm{max}}) \big) \\
     -\tan(\delta_{\mathrm{max}}) &  1 &  \big(l_{\mathrm{f}} - w_{\mathrm{r}} \tan(\delta_{\mathrm{max}}) \big) \\
      \tan(\delta_{\mathrm{min}}) & -1 & -\big(l_{\mathrm{f}} + w_{\mathrm{l}} \tan(\delta_{\mathrm{min}}) \big) \\
      \tan(\delta_{\mathrm{min}}) & -1 &  \big(l_{\mathrm{r}} - w_{\mathrm{l}} \tan(\delta_{\mathrm{min}}) \big) \\
      \tan(\delta_{\mathrm{min}}) & -1 &  \big(l_{\mathrm{r}} + w_{\mathrm{r}} \tan(\delta_{\mathrm{min}}) \big) \\
      \tan(\delta_{\mathrm{min}}) & -1 & -\big(l_{\mathrm{f}} - w_{\mathrm{r}} \tan(\delta_{\mathrm{min}}) \big) \\
  \end{bmatrix},
  \label{eq:constraintmatrixB}
\end{align}

\noindent with
constraints ${v_x-w_{\mathrm{l}}\dot{\psi}} < 0$ for wheels $1$ and $2$, and
${v_x+w_{\mathrm{r}}\dot{\psi}} < 0$ for wheels $3$ and $4$. In
\eqref{constraintmatrixB} rows $i$ and $i+4$ come from wheel $i$, vectors that are represented by
$\bs{b}^{(1)}_i, \bs{b}^{(2)}_i$ in \eqref{b_i} and correspond to the conditions
$\delta_i = \delta_{\mathrm{max}}$ and $\delta_i = \delta_{\mathrm{min}}$ respectively,
where $i \in \{1, 2, 3, 4\}$.
Note that the additional
conditions ${v_x-w_{\mathrm{l}}\dot{\psi}} < 0$
and ${v_x+w_{\mathrm{r}}\dot{\psi}} < 0$ prevent the configuration space to be over-constrained,
that is $x<0$ for $\operatorname{atan2}(y, x) \in (\frac{\pi}{2},\pi)$.
It is also worth noting that the first four rows of \eqref{constraintmatrixB} are equivalent to
the expressions in the first row of \eqref{b_i}; the negative signs
are used solely to
distinguish between the conditions $\delta_i = \delta_{\mathrm{max}}$ and
$\delta_i = \delta_{\mathrm{min}}$.

There is only one large region, which is the most desirable one to keep the configuration vector
in. Note that the existence of many small regions associated with $v_x<0$ will cause the wheels
to flip often if the robot is moving in reverse direction. To keep the number of wheel flips
low, backward motion should be kept within the second largest region (behind all
discontinuity planes). We specify these two regions as the preferred regions
and make it the motion planner's responsibility to generate the velocities within them.

\subsection{Region Signature}
\label{sec:regsig}

A generic way to determine the region for a given velocity vector
$\bs{v}$ is to calculate
\begin{align}
  \bs{s}=\frac{1}{2}\Big(\operatorname{\mathbf{sign}}\big(\mtrx{B}\bs{v}\big) + \mathbf{1}_{\mathrm{8 \times 1}}\Big),
  \label{eq:calcRegionID}
\end{align}

\noindent where $\operatorname{\mathbf{sign}}(.)$ is the sign function applied
to a vector component-wise.

\begin{figure}[htp!]
  \vspace{1mm}
  \centering
  \includegraphics[width=0.6\linewidth]{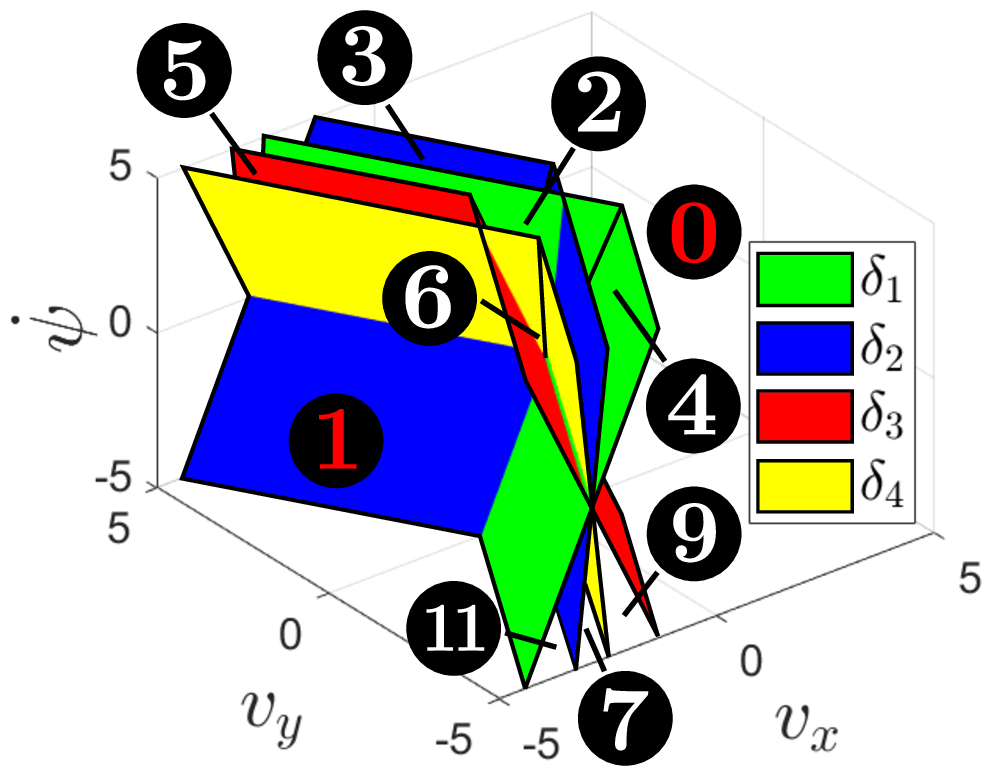}
  \caption{Definition of 12 continuous regions in the velocity configuration space.
    Note that regions with ID 8 and 10 are hidden at the bottom of the left halves of the planes. Regions with IDs 0 and 1 (red) are preferred.}
  \label{fig:velocityRegions}
\end{figure}

The resulting vector $\bs{s}$, which we call the
{\em region signature},
indicates the region in which the velocity vector resides.
Specifically, $s_j=1$ indicates that $\bs{v}$ is ahead of the plane $j$
defined by the constraint of the $j$th row of $\mtrx{B}$.
Vector $\bs{s}$ can be interpreted as binary-encoded unsigned integer
and used as an index into a lookup table that retrieves the ID of one
of the 12 regions partitioned by the 8 discontinuity planes. The numerical
values of the region IDs are visualized in \figref{velocityRegions}.
The solution space for \eqref{calcRegionID} does not span the full
8-bit integer space. Instead, there are only 46 possible
solutions, which in turn map to 12 different regions (multiple
solutions correspond to the same region).

Region 0 is the largest continuous
region shown in \figref{2pi3} and \figref{3pi4} and is typically the preferred region of the
configuration space. It corresponds to the robot moving primarily forward and sideways. Region 1
is the second largest region and corresponds to the generally backwards motion. Other
regions are not preferred because they are small and thus likely to trigger the
transition through discontinuity planes, which requires the robot to stop and reposition
the wheels.

The design objective of the motion planner is to generate a time-series
of velocities that advances the robot towards the goal, steer clear
of obstacles and avoid crossing the discontinuity planes whenever possible.
The last requirement is not a hard constraint but rather a strong
preference. The planner should do the best effort to keep the generated
velocity in region 0 (preferred region) or region 1 (second-preferred region),
and cross into other (non-preferred) regions only when
forced by other constraints. Once in a non-preferred region, the
planner can trade off between restricting itself to the new
and smaller region or taking the penalty to transition back to the
preferred region where it has more options for selecting the velocity
vectors. In either case, the region signature of the present and the previous
velocity can be used as an input to a cost function that controls
the selection of the velocity vector.

\section{Planner Implementation}
\label{sec:planner}

The principles presented thus far are valid for any navigation stack
that follows the cascade architecture. The constraints that we
derived belong to the local planner, which generates velocity
commands based on the waypoint the robot is trying to reach and the
cost map constructed from observed obstacles.
The motion controller, which
converts the velocity vector to actuation defends from discontinuity-plane
crossing by stopping the motion if the wheel flip becomes necessary.

To evaluate our system in simulation and on a physical robot,
we chose to implement the planner as an extension to the
DWA\cite{fox1997dynamic} planner, which comes with the standard
ROS Navigation\cite{ros_navigation} stack and gives us a
well understood baseline to compare the results against.

\subsection{DWA Planner}
The ``stock'' DWA planner explores the space of velocities
that are reachable from present velocity given robot dynamics.
For each candidate velocity, ($v_x$, $v_y$, and $\dot{\psi}$),
DWA forward-simulates
the near-term robot position and passes the projected paths
to the chain of critic functions.
Each critic computes the cost (or flags the velocity inadmissible)
independently. The selected velocity is the admissible velocity
with the lowest weighted sum produced by each critic.
Our implementation adds two new critics to the chain
called {\em swerve-constraint critic}, and
{\em smoothness critic}.

\subsection{Swerve-Constraint Critic}
The swerve-constraint critic penalizes
velocities in non-preferred regions and velocities
that cross from one region to the other. This requires access to both
proposed velocity and present velocity. The former is generated
by the planner, whereas the latter can either come from odometry
feedback or the velocity command of the previous planning cycle.

\begin{figure}[bht!]
  \vspace{1mm}
  \centering
  \subfigure[]{
		\centering
		\label{fig:distanceToPlaneA}
		\includegraphics[width=0.47\linewidth]{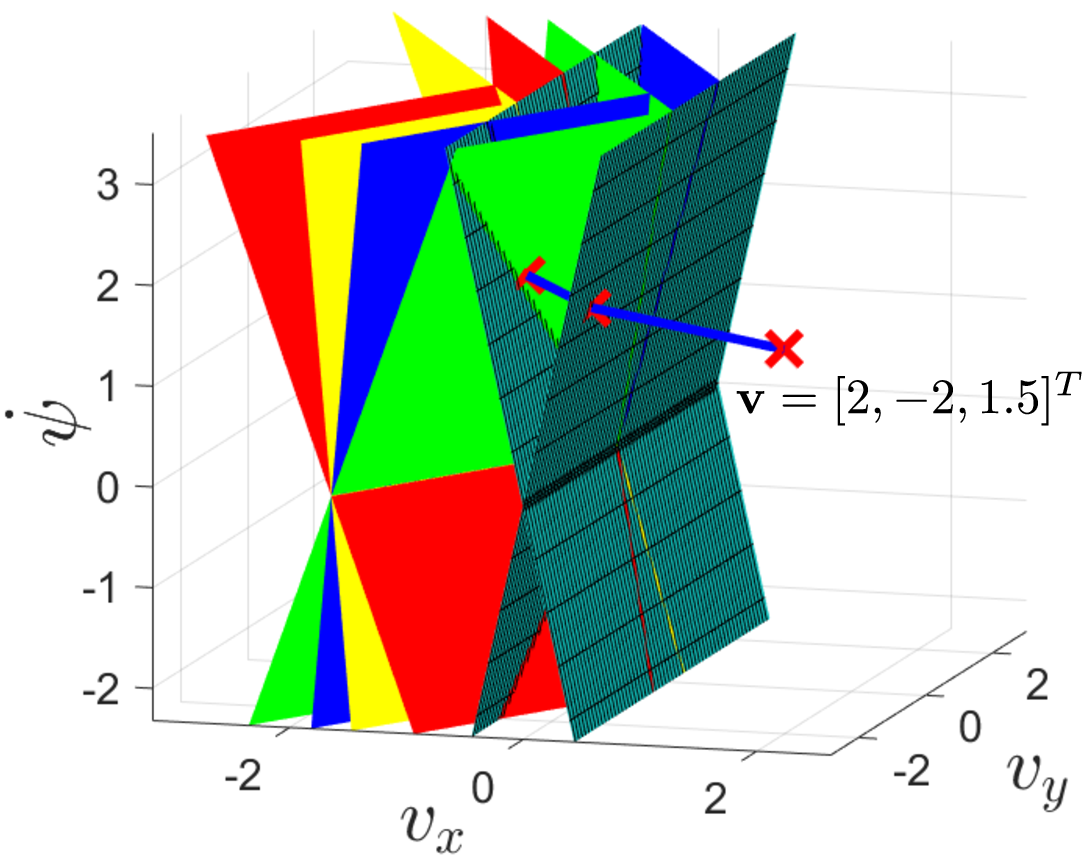}
	}
  \subfigure[]{
		\centering
		\label{fig:distanceToPlaneB}
		\includegraphics[width=0.45\linewidth]{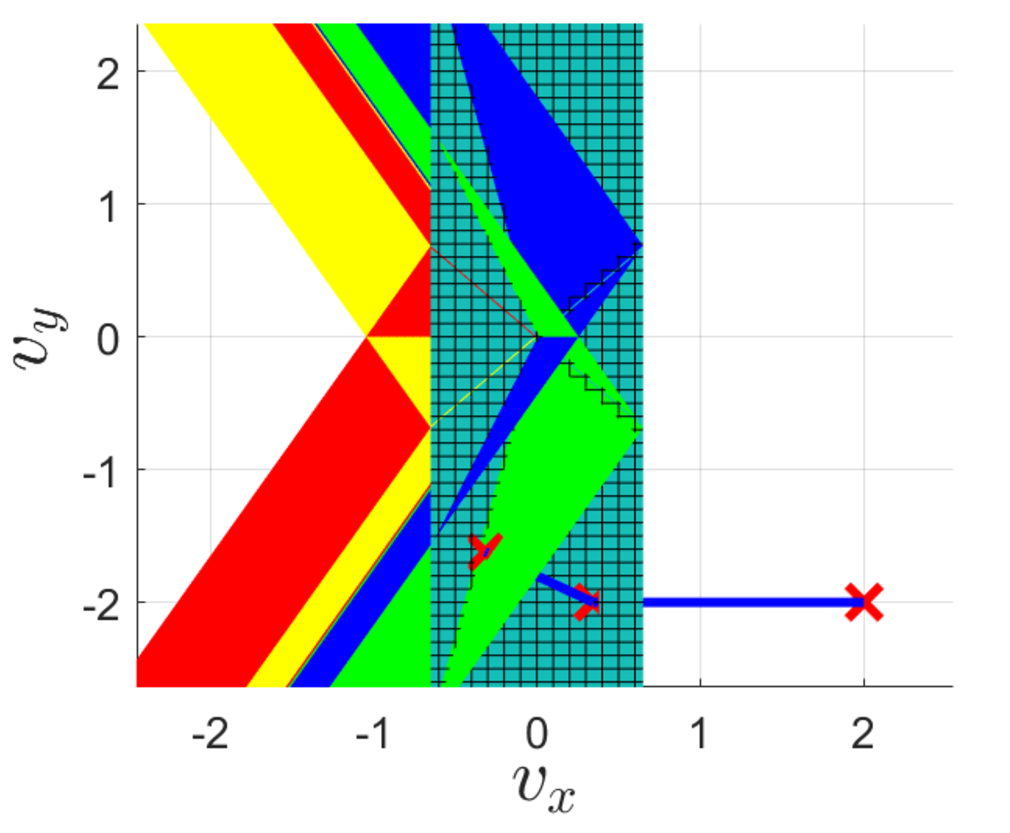}
	}
  \caption{Distance to the nearest discontinuity plane for a velocity vector located in region 0 and ahead of
    two auxiliary planes. (a) side view and (b) top view. Red markers show
    the original velocity vector, its projection onto the nearest auxiliary plane, followed by
    the projection onto the nearest discontinuity plane. The blue lines represent
    the resulting distance metric.}
  \label{fig:distanceToPlane}
\end{figure}

We implement two scoring modes.
In {\em simple scoring} mode, the critic first checks if the proposed velocity
is in the same region as the present one. If so,
it assigns zero-cost if the region is preferred, and half of the maximum cost,
otherwise. Transitions into a non-preferred region are deemed inadmissible
and all other transitions are penalized with the maximum cost.
In {\em distance-based scoring} mode, transitions within the same region
are penalized
using exponentially decaying function of the distance to the nearest
discontinuity plane (denoted by $d_{\min}$), with the cost given by
\begin{equation}
  c_{\text{swerve}} = \alpha_{\text{swerve}} e^{-\gamma d_{\min}}
  \label{eq:costswerve}
\end{equation}
where $\alpha_{\text{swerve}}$ is the maximum cost for penalizing
the violation of swerve constraints and $\gamma$
is the decaying rate.
This scoring policy reduces the probability of region crossing by keeping the velocity vector
away from the discontinuity plane.
The distance for velocity located in non-preferred regions is
simply the minimum normal distance to all the discontinuity planes:

\begin{equation}
  d_{\min} = \min_{j=[1,8]}\{d_j\}, \: \text{ with } d_j = \frac{\mtrx{B}[j,:]\bs{v}}{\|\mtrx{B}[j,:]\|},
  \label{eq:distancetoplane}
\end{equation}

\noindent where $j \in [1, 8]$ is the plane index, $\mtrx{B}[j,:]$ is the $j$th
row of the matrix $\mtrx{B}$, and $\|.\|$ is the 2-norm of a vector.
For region 1, the distance is the minimum
distance to four discontinuity planes that form the region, namely the left halves of
the planes for $\delta_2$ and $\delta_4$, and the right halves for $\delta_1$ and $\delta_3$.
Region 0 is non-convex and the distance defined by \eqref{distancetoplane}
does not exist, so we use a modified metric.
We first evaluate the velocity vector against two auxiliary planes
${v_x-w_{\mathrm{l}}\dot{\psi}}=0$ and ${v_x+w_{\mathrm{r}}\dot{\psi}}=0$.
If the velocity is located behind either of two auxiliary planes,
the distance is still that of \eqref{distancetoplane}.
If the velocity vector is located ahead of these two auxiliary
planes, we first find the projection of the velocity vector onto the nearest auxiliary
plane and compute the distance,
followed by adding the distance from that projection to
the nearest constraint-plane (see \figref{distanceToPlane}).

The exception to the above-described policy is when the robot is starting
the motion from a stationary or close-to-stationary state.
In that case, the critic allows
the velocity vector to transition to any region, because there is
no notion of discontinuity. In practice, this allows the robot
to use non-preferred regions when they are really needed, but as the
motion starts, the configuration will gravitate towards preferred
regions. Once in full motion, preferred regions will be used barring
external disturbances, actuator errors, or tuning that makes other
critics in the chain prevalent. If a non-preferred region is entered,
the motion planner will try to transition the velocity configuration back into the
preferred region as soon as such transition becomes possible.

Computational load per proposed velocity is relatively small. The critic
must do 8 matrix multiplications of \eqref{distancetoplane}, where $\mtrx{B}$ is
static and can be pre-calculated at initialization time. The dimension of
$\mtrx{B}$ is $8\times3$, which is considered small and the
final cost \eqref{costswerve} needs to be calculated only once for each proposed
velocity.

\subsection{Smoothness Critic}

As the DWA planner executes the chain of critics, it may assign
the same cost to multiple candidate velocities
that are located far apart in the configuration space. Such scenarios
commonly occur when navigating around obstacles or in narrow
spaces where only a portion of the sampled velocities are admissible.

If a velocity previously commanded is deemed
inadmissible in the next planning cycle, the planner might select another
candidate velocity that may not necessarily trigger the discontinuity crossing,
but still cause a large step change in wheel position.
This can be harmful if wheels are controlled independently.
Specifically, if one wheel gets ahead of the other
the actuation may no longer correspond to any valid velocity configuration,
which causes lateral friction.

A simple way to counter the problem is to make the planner
prefer gradual transitions. The smoothness critic
computes the distance to the previously commanded velocity and penalizes
the proposed velocity linearly with the calculated distance. The cost of smoothness
critic is computed as
\begin{equation}
  c_{\text{smoothness}} = \alpha_{\text{smoothness}} \mathrm{cap}(\bs{v}^d, \bs{v}^c)
\end{equation}
with $\alpha_{\text{smoothness}}$ parametrizing the maximum cost of the smoothness critic,
$\bs{v}^d$ and $\bs{v}^c$ being the proposed and the present velocity. The $\mathrm{cap}(.,.)$
function is  defined as
\begin{align}
  \mathrm{cap}(\bs{v}^d, \bs{v}^c) = \left\{
    \begin{array}{ll}
      \frac{\|\bs{v}^d-\bs{v}^c\|}{\Delta v_{\max}}, & \text{if} \: \|\bs{v}^d-\bs{v}^c\| \le \Delta v_{\max}, \\
      1, & \: \text{otherwise},
    \end{array}\right.
\end{align}
where $\|.\|$ is the 2-norm of a vector
and $\Delta v_{\max}$ determines the smallest velocity transition step
penalized with the maximum cost.

Computational load per
proposed velocity is trivial
because the critic only needs to subtract the current and proposed
velocity and compare the magnitude of the difference with a threshold.

\section{Experimental Results}
\label{sec:results}

We evaluate the performance both on physical hardware and in simulation
using the Gazebo\cite{Koenig2004} package for high-fidelity physics modeling.
The models capture full dynamics of the robot frame, motor dynamics,
and contact points between wheels and the surface. Sensor noise and
non-deterministic effects of wheel-surface interaction are also captured
by the model and we do not
reset the simulation world between runs. Consequently, stochastic effects
seen in the physical world are also present in simulation.

Physical robot build has the wheel base $l_\mathrm{f}=l_\mathrm{r}=w_\mathrm{l}=w_\mathrm{r}=0.2m$ and steering limits $\delta_\mathrm{max}=-\delta_\mathrm{min}=130^{o}$. It is driven by brushless-motor hub-wheels of radius
$0.08m$ and steered using a brushed DC motor. The motion-control/actuation
loop runs at 100Hz, the local planner runs at 5Hz, and the global planner
re-plans once per second.

The parameters for swerve-constraint and
smoothness critics are set to the same values in both simulation
and real-world experiments, which are $\alpha_{\text{swerve}}=5.0$,
$\alpha_{\text{smoothness}}=2.0$, $\gamma=20.0$, and $\Delta v_{\max}=0.2$.
The maximum cost value for the swerve-constraint critic was selected such that
its weight is about 60\% of the weight of (standard DWA) path-distance and
goal-distance critics. In this way, progress towards the goal and remaining
on the planned path is the priority, but motions that are likely to cause
discontinuity are discouraged.  The smoothness critic must have the smallest
weight relative to other critics so that it does not interfere with the ability
to follow the path and avoid obstacles. Fine tuning of these weights is done
experimentally. Parameters $\gamma$
and $\Delta v_{\max}$ were preliminary determined through simulation to ensure efficiency,
specifically by providing an appropriate decay rate and a representative velocity difference
for characterizing non-smoothness.

\subsection{Experiments}

For each test case, we issue a sequence of goals
that make the robot follow an established
nominal path. For each run,
we measure the time to complete the trajectory
and count the wheel-flip events due to discontinuity-plane crossings.
For both metrics, the lower value indicates better performance.

The paths we use include simple patterns, such as rectangle,
figure-x and figure-8 path, as well as a
more complex path in which the robot navigates through a maze.
The maze path
consists of segments which expose the robot to a
variety of maneuvers, such as turns, change in direction, rotation,
sideways motion, and backwards motion.
The nominal path for each scenario is the same, but environment
variations may produce slightly different waypoint coordinates for
different runs, which in turn impacts the discontinuity results.
For each scenario, we repeat the test $10$ times to ensure that the
results are repeatable and to quantify the robot behavior variability.

We evaluate the planner in simple-scoring mode and distance-based
scoring mode. For both modes, we evaluate the case when only
region 0 is preferred (referred to as the forward-only critic)
and the case where both region 0 and 1 are preferred
(referred to as the forward-backward critic).
All other regions are non-preferred.

We compare the performance of our augmented DWA planner
with two baseline cases: 1) ``stock'' DWA planner
with basic motion controller, and 2) ``stock'' DWA planner with motion
controller that picks the wheel orientation that results in the
shortest angle transition if the kinematic solution yields two
valid orientations. The latter helps with the stock planner,
but can interfere with the augmented planner
because the controller-side optimization can transition the wheel
into the region that is inconsistent with planner's belief.

\subsection{Results}

The planner configurations are detailed in \tabref{config}, and
\tabref{sim_results} (simulation) and \tabref{exp_results} (physical robot)
summarize the number of encountered discontinuities and travel time.
We highlight the ``winner'' configuration according to each metric, which
is selected by prioritizing a lower median value, alongside a reasonably
low mean and a lower standard deviation, to ensure both typical performance
and consistency across repeated runs.

\begin{table}[ht!]
  \caption{Configurations of the evaluated planners.}
  \centering
  \begin{tabular}{l|c|c|c}
    \hline\hline
    \multirow{2}{*}{\textbf{Config. ID}} & \textbf{Shortest} & \multirow{2}{*}{\textbf{Scoring mode}} & \multirow{2}{*}{\textbf{Preferred regions}} \\
    & \textbf{transition} &  & \\
    \hline
    baseline-1 & no & - & - \\
    baseline-2 & yes & - & - \\
    \hline
    aug-DWA-1 & no & distance-based & forward-only \\
    aug-DWA-2 & no & distance-based & forward-backward \\
    aug-DWA-3 & no & simple-scoring & forward-only \\
    aug-DWA-4 & no & simple-scoring & forward-backward \\
    \hline\hline
  \end{tabular}
  \label{tab:config}
\end{table}

\begin{table}[ht!]
  \caption{Simulation results: discontinuity count and travel time.}
  \scriptsize
  \centering
  \begin{tabular}{c|l|c|c|c|c|c|c}
    \hline\hline
    \multirow{2}{*}{\textbf{Scenario}} & \multirow{2}{*}{\textbf{Config. ID}} &
    \multicolumn{3}{c|}{\textbf{Discontinuity count}} & \multicolumn{3}{|c}{\textbf{Travel time (s)}} \\
    \cline{3-8} & & \textbf{\tiny{mean}} & \textbf{\tiny{median}} & \textbf{\tiny{std}} & \textbf{\tiny{mean}} & \textbf{\tiny{median}} & \textbf{\tiny{std}} \\
    \hline\hline
    \multirow{6}{*}{figure-8} & baseline-1 & 11.1 & 7.5 & 10.6 & 52.0 & 44.3 & 15.7 \\
    & baseline-2 & 3.7 & 3.0 & 2.7 & 39.7 & 39.5 & 2.3 \\
    \cline{2-8} & aug-DWA-1 & 1.2 & 1.0 & 1.2 & 38.4 & 37.1 & 2.6 \\
    & aug-DWA-2 & \textbf{1.2} & \textbf{1.0} & \textbf{0.6} & 38.5 & 38.3 & 1.3 \\
    & aug-DWA-3 & 1.6 & 1.5 & 0.7 & \textbf{37.3} & \textbf{37.3} & \textbf{1.4} \\
    & aug-DWA-4 & 1.5 & 1.0 & 1.0 & 38.4 & 38.4 & 1.6 \\
    \hline\hline
    \multirow{6}{*}{figure-x} & baseline-1 & 30.9 & 26.5 & 9.9 & 89.4 & 84.9 & 12.8 \\
    & baseline-2 & 5.9 & 6.5 & 3.4 & 62.3 & 61.6 & 5.4 \\
    \cline{2-8} & aug-DWA-1 & 4.2 & 4.0 & 2.0 & 59.3 & 58.2 & 5.8 \\
    & aug-DWA-2 & \textbf{2.2} & \textbf{2.0} & \textbf{1.9} & \textbf{56.0} & \textbf{56.0} & \textbf{6.5} \\
    & aug-DWA-3 & 5.8 & 6.0 & 1.6 & 64.4 & 67.2 & 7.3 \\
    & aug-DWA-4 & 3.1 & 4.0 & 2.2 & 59.1 & 58.3 & 8.7 \\
    \hline\hline
    \multirow{6}{*}{rectangle} & baseline-1 & 15.4 & 8.5 & 14.6 & 53.9 & 47.9 & 16.8 \\
    & baseline-2 & 5.5 & 6.0 & 2.2 & 42.7 & 43.7 & 3.4 \\
    \cline{2-8} & aug-DWA-1 & \textbf{1.5} & \textbf{1.0} & \textbf{1.0} & 39.5 & 39.5 & 1.9 \\
    & aug-DWA-2 & 2.2 & 2.5 & 1.2 & 39.4 & 38.9 & 1.7 \\
    & aug-DWA-3 & 3.3 & 3.5 & 2.7 & 41.9 & 40.8 & 4.2 \\
    & aug-DWA-4 & 2.6 & 2.5 & 1.4 & \textbf{39.0} & \textbf{38.6} & \textbf{1.8} \\
    \hline\hline
    \multirow{6}{*}{maze} & baseline-1 & 18.3 & 11.5 & 19.3 & 142.1 & 132.9 & 23.8 \\
    & baseline-2 & 7.3 & 8.0 & 3.7 & 128.1 & 128.2 & 3.2 \\
    \cline{2-8} & aug-DWA-1 & 3.3 & 3.0 & 1.0 & 123.3 & 122.8 & 2.1 \\
    & aug-DWA-2 & \textbf{2.3} & \textbf{2.5} & \textbf{1.2} & 124.1 & 123.1 & 3.7 \\
    & aug-DWA-3 & 4.2 & 4.0 & 1.9 & 125.6 & 125.8 & 2.9 \\
    & aug-DWA-4 & 2.6 & 2.5 & 1.7 & \textbf{123.1} & \textbf{122.7} & \textbf{2.1} \\
    \hline\hline
  \end{tabular}
  \label{tab:sim_results}
\end{table}

\begin{table}[ht!]
  \caption{Physical-robot results: discontinuity count and travel time.}
  \scriptsize
  \centering
  \begin{tabular}{c|l|c|c|c|c|c|c}
    \hline\hline
    \multirow{2}{*}{\textbf{Scenario}} & \multirow{2}{*}{\textbf{Config. ID}} &
    \multicolumn{3}{c|}{\textbf{Discontinuity count}} & \multicolumn{3}{|c}{\textbf{Travel time (s)}} \\
    \cline{3-8} & & \textbf{\tiny{mean}} & \textbf{\tiny{median}} & \textbf{\tiny{std}} & \textbf{\tiny{mean}} & \textbf{\tiny{median}} & \textbf{\tiny{std}} \\
    \hline\hline
    \multirow{6}{*}{figure-8} & baseline-1 & 18.6 & 15.5 & 14.2 & 166.3 & 172.0 & 85.6 \\
    & baseline-2 & 4.8 & 5.0 & 3.3 & 76.9 & 76.7 & 24.4 \\
    \cline{2-8} & aug-DWA-1 & 1.4 & 1.0 & 2.2 & \textbf{46.4} & \textbf{40.3} & \textbf{13.5} \\
    & aug-DWA-2 & 2.0 & 2.0 & 1.8 & 50.1 & 48.8 & 10.2 \\
    & aug-DWA-3 & 2.2 & 1.5 & 2.5 & 52.8 & 47.3 & 18.3 \\
    & aug-DWA-4 & \textbf{1.5} & \textbf{1.0} & \textbf{1.6} & 48.4 & 49.3 & 12.6 \\
    \hline\hline
    \multirow{6}{*}{figure-x} & baseline-1 & 38.4 & 36.0 & 13.7 & 319.6 & 327.2 & 81.5\\
    & baseline-2 & 8.4 & 9.0 & 3.2 & 153.9 & 164.2 & 52.2 \\
    \cline{2-8} & aug-DWA-1 & 6.8 & 7.0 & 1.6 & 87.3 & 88.3 & 6.7 \\
    & aug-DWA-2 & \textbf{4.3} & \textbf{4.0} & \textbf{2.5} & \textbf{72.4} & \textbf{70.2} & \textbf{10.1} \\
    & aug-DWA-3 & 8.4 & 6.5 & 4.2 & 97.2 & 100.9 & 21.8 \\
    & aug-DWA-4 & 5.6 & 5.5 & 2.4 & 90.8 & 88.1 & 15.2 \\
    \hline\hline
    \multirow{6}{*}{rectangle} & baseline-1 & 16.2 & 12.0 & 13.9 & 135.4 & 125.1 & 69.1 \\
    & baseline-2 & 2.2 & 2.0 & 0.92 & 53.9 & 53.6 & 9.0 \\
    \cline{2-8} & aug-DWA-1 & 1.7 & 1.5 & 1.2 & 48.1 & 48.1 & 7.7 \\
    & aug-DWA-2 & 1.8 & 2.0 & 1.3 & 50.5 & 51.3 & 6.9 \\
    & aug-DWA-3 & \textbf{1.7} & \textbf{1.0} & \textbf{1.4} & \textbf{47.6} & \textbf{44.5} & \textbf{8.9} \\
    & aug-DWA-4 & 2.0 & 2.0 & 1.1 & 53.7 & 52.5 & 9.1 \\
    \hline\hline
    \multirow{6}{*}{maze} & baseline-1* & 35.3 & 16.0 & 41.5 & 337.0 & 204.2 & 265.1 \\
    & baseline-2 & 8.8 & 10 & 2.8 & 215.2 & 194.8 & 61.4 \\
    \cline{2-8} & aug-DWA-1 & 3.7 & 3.0 & 1.6 & 149.3 & 147.1 & 9.9 \\
    & aug-DWA-2 & \textbf{2.2} & \textbf{2.0} & \textbf{1.0} & \textbf{137.5} & \textbf{137.8} & \textbf{3.7} \\
    & aug-DWA-3 & 4.1 & 3.0 & 2.8 & 149.0 & 144.8 & 14.0 \\
    & aug-DWA-4 & 2.5 & 2.0 & 2.0 & 141.5 & 139.4 & 8.1 \\
    \hline\hline
    \multicolumn{8}{l}{*Baseline-1 maze-test abandoned due to incessant wheel thrashing.}\\
  \end{tabular}
  \vspace{-5mm}
  \label{tab:exp_results}
\end{table}

Physical robot experiments show the same general trends as in the simulation,
so in that sense they are consistent. The difference in absolute numbers
is expected, because in simulation there is no room clutter that interferes
with the localization system, wheel-surface interaction is more predictable,
and the robot can recover from discontinuity crossings more easily and quickly.

The baseline planner performance appears very poor compared
to all other configurations. In some runs
the robot remained stuck
thrashing the wheels, because the planner kept issuing
the discontinuity-crossing velocities. This situation happens
when the robot needs to move in the direction of the diagonal formed
by its center and one of its rear edge, putting the wheels at the
boundary of the discontinuity plane.
For this reason we had to abandon the physical-robot test for the baseline-1
configuration in the maze scenario.
Although the shortest-transition enhancement in baseline-2 improved the stock
DWA planner considerably, the advantage of using the augmented DWA
planner are still evident across all scenarios and configurations.
Specifically, in challenging cases (e.g., figure-x scenario),
baseline-2 achieves optimization with respect to discontinuity crossings, but
overall travel time is significantly longer than that of the augmented planner
(i.e., aug-DWA-3). The travel-time improvement arises from
our optimization that enables the robot to operate away from inefficient
regions.

The distance-based scoring mode performs generally
better than the simple-scoring mode. The performance difference for
exceptions (figure-8 and rectangle scenarios on the physical robot) is statistically
insignificant.
% Further, the travel-time improvement appears smaller in
% simulation, because the wheel flipping is faster in the simulation model.
Performance difference
between forward-only and forward-backward configurations
depends on whether the natural travel pattern is amenable to
driving in reverse.
Specifically, figure-8 and rectangle scenarios seem to prefer the
forward-only configuration, whereas the figure-x scenario prefers
bidirectional motion. This is expected, because when the next waypoint
is behind the robot, the forward only configuration will tend to produce
a U-turn with forward-moving velocity, which is a motion pattern likely
to trigger the discontinuity crossings.

\begin{figure}[htp!]
  \centering
  \includegraphics[width=0.8\linewidth]{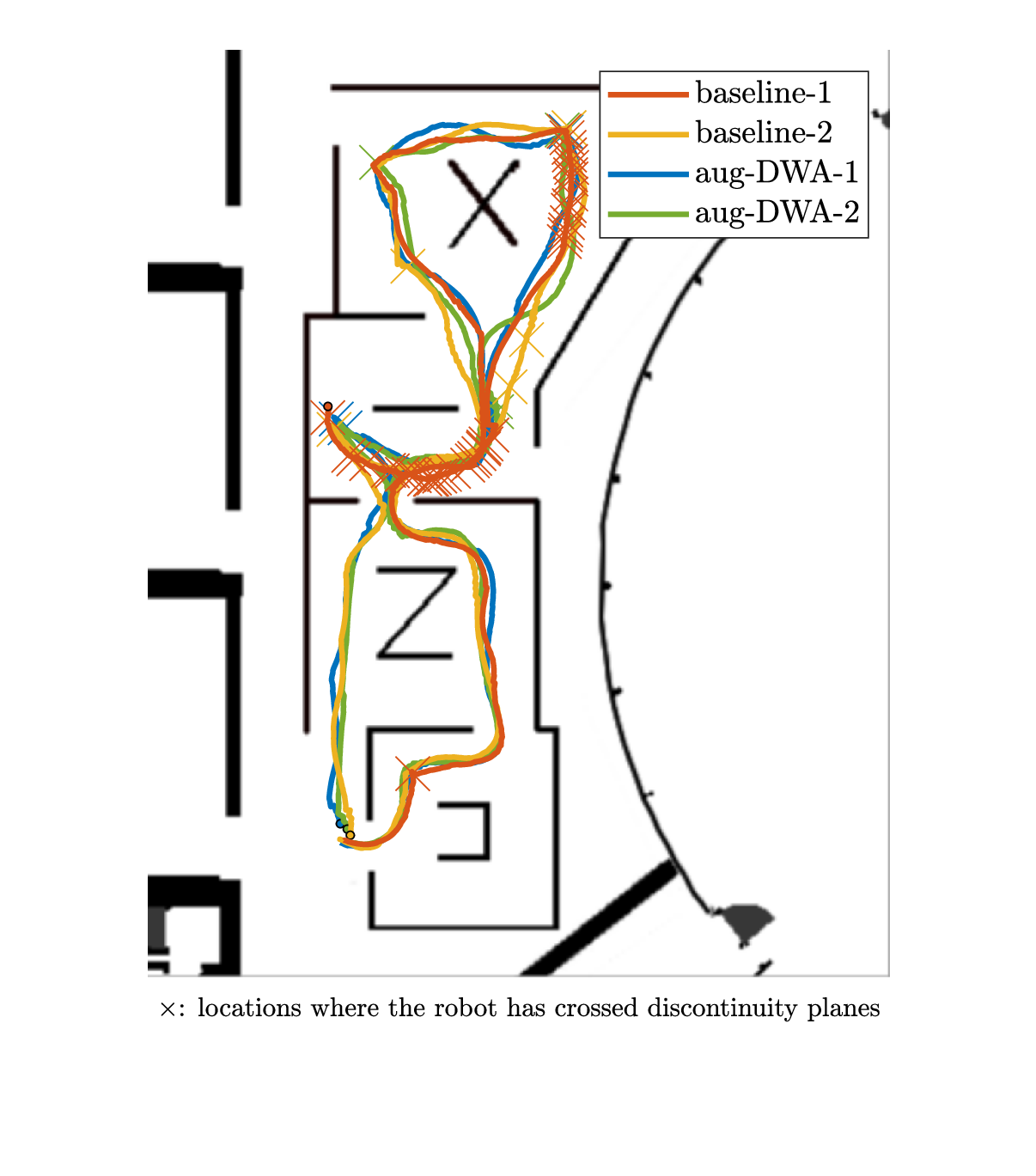}
  \vspace{-1cm}
  \caption{Actual robot trajectories with two baseline cases and two instances of the augmented DWA planner proposed in this paper.
    Discontinuity counts: baseline-1: 39 (aborted), baseline-2: 5, aug-DWA-1: 3, aug-DWA-2: 4.}
  \label{fig:robotTraj}
\end{figure}

\figref{robotTraj} shows actual trajectories followed recorded from the
physical-robot experiment running in the maze scenario. Traversed paths
are well aligned, which means that all configurations are tracking the
waypoints produced by the global planner well. Minor
variations come from the variations in the global planner output and
path tracking performance.

The main performance difference comes from the number of
discontinuity crossings depicted by $\times$-markers in the figure.
Baseline configurations are dominated by discontinuity crossings
in the upper section of the maze because that is where the
robot tends to change the direction of travel or attempts
to move sideways. The last segment of the baseline-1 case (red line)
is missing because the goal was abandoned.
As expected, the number of discontinuity crossings
has been dramatically reduced using two instances of the augmented DWA
planner.

\section{Conclusions}
\label{sec:conclusions}

Steering constraints of a 4WIS drive introduce discontinuity
planes in the 3-dimensional velocity configuration space. Previously
reported results have only studied unconstrained steering or
side-stepped the constrained-steering problem by using Ackermann
steering. To our knowledge, our work is the first that addresses
the constrained-steering problem for 4WIS by deriving the discontinuity
planes, followed
by designing an example planner that uses them as an integral
part of decision making.
% Discontinuity planes are general kinematic constraints, which
% dissolve when the mechanical design allows full 360-degree swerve
% and reduce to Ackermann steering for steering limits below 90 degrees.

In our implementation, the local planner scores proposed velocities based on
the distance to discontinuity planes and the motion controller stops
the motion and flips the wheel if necessary. The planner prefers
the velocities that avoid the wheel flip, but it does not
eliminate them. The overall behavior depends on other critics
in the scoring chains that are used and the relative weights among
the critics. We further introduced an additional smoothness critic
to cope with
the wheel-racing condition that arises if each wheel is controlled
with an independent PID controller. Coupled wheel control is the logical
follow-up to the presented work and so is a fully dynamic controller.
The constraints we derived herein remain valid in both cases.

\section*{Acknowledgments and Notes}

We thank Liyu Cai, Haiyang Zhang, Yonggang Wang, and Li Yang who built the
robot mechanical frame that we used for experiments.
Anthony Rodrigues, Andrew Hollabaugh, and Joshua Schmidt implemented
hardware and software improvements that enabled the implementation of the
presented controller. Larry O'Gorman and Martin Carroll provided invaluable
comments that helped us improve the paper.

Extended video demonstration of the results is available
at \url{https://youtu.be/8l9s2Wb_vec.}

% IEEE
\begin{small}
\bibliographystyle{IEEEtran}
\bibliography{ArabRef}
\end{small}

\end{document}